\documentclass{article}

\usepackage[preprint]{neurips_2023}

\usepackage{natbib}
\setcitestyle{numbers,square}
\usepackage[utf8]{inputenc} 
\usepackage[T1]{fontenc}    
\usepackage{hyperref}       
\usepackage{url}            
\usepackage{booktabs}       
\usepackage{amsfonts}       
\usepackage{nicefrac}       
\usepackage{microtype}      
\usepackage{xcolor}         
\usepackage{booktabs}
\usepackage{multirow}
\usepackage{graphicx}

\title{Pushing the Limits of 3D Shape Generation at Scale}

\author{%
  Yu Wang\textsuperscript{1,}\thanks{Equal contribution.}\;\;\;
  Xuelin Qian\textsuperscript{1,}\footnotemark[1]\;\;\;
  Jingyang Huo\textsuperscript{1}\;\;\;
  Tiejun Huang\textsuperscript{2}\;\;\;
  Bo Zhao\textsuperscript{2,}\thanks{Corresponding authors.}\;\;\;
  Yanwei Fu\textsuperscript{1,}\footnotemark[2]\\
    \textsuperscript{1}Fudan University\quad\quad \textsuperscript{2}Beijing Academy of Artificial Intelligence\\
  \texttt{\{yu\_w13,xlqian,yanweifu\}@fudan.edu.cn\;\;\;zhaobo@baai.ac.cn} \\
}

\begin{document}

\maketitle

\begin{abstract}

We present a significant breakthrough in 3D shape generation by scaling it to unprecedented dimensions. Through the adaptation of the Auto-Regressive model and the utilization of large language models, we have developed a remarkable model with an astounding 3.6 billion trainable parameters, establishing it as the largest 3D shape generation model to date, named \emph{Argus-3D}. Our approach addresses the limitations of existing methods by enhancing the quality and diversity of generated 3D shapes.
To tackle the challenges of high-resolution 3D shape generation, our model incorporates tri-plane features as latent representations, effectively reducing computational complexity. Additionally, we introduce a discrete codebook for efficient quantization of these representations. Leveraging the power of transformers, we enable multi-modal conditional generation, facilitating the production of diverse and visually impressive 3D shapes.
To train our expansive model, we leverage an ensemble of publicly-available 3D datasets, consisting of a comprehensive collection of approximately 900,000 objects from renowned repositories such as ModelNet40, ShapeNet, Pix3D, 3D-Future, and Objaverse. This diverse dataset empowers our model to learn from a wide range of object variations, bolstering its ability to generate high-quality and diverse 3D shapes.
Through extensive experimentation, we demonstrate the remarkable efficacy of our approach in significantly improving the visual quality of generated 3D shapes. By pushing the boundaries of 3D generation, introducing novel methods for latent representation learning, and harnessing the power of transformers for multi-modal conditional generation, our contributions pave the way for substantial advancements in the field. Our work unlocks new possibilities for applications in gaming, virtual reality, product design, and other domains that demand high-quality and diverse 3D objects. Project page and code: \href{https://argus-3d.github.io}{https://argus-3d.github.io}.

\end{abstract}

\section{Introduction}
\label{section1}

The field of 3D shape generation has gained significant attention due to its wide range of applications, and it continues to be an active area of research. Existing methods predominantly rely on generative adversarial networks (GANs) \cite{3dgan, pcgan, abdal20233davatargan, graf, pigan, giraffe, chan2022efficient, or2022stylesdf, gu2021stylenerf, zhou2021cips, pavllo2021learning, hao2021gancraft, wen2021learning, chen2021decor, luo2021surfgen, li2021sp, ibing20213d}, variational autoencoders (VAEs) \cite{mo2019structurenet, gao2019sdm, gao2021tm, kim2021setvae}, flows \cite{pointflow, sanghi2022clip, kim2020softflow, klokov2020discrete}, autoregressive models \cite{pointgrow, polygen, ibing2021octree, iccv}, denoising diffusion probabilistic models (DDPMs) \cite{lion, cheng2023sdfusion, rodin, luo2021diffusion, zhou20213d,meshdiffusion,Diffusion-Based,zhang20233dshape2vecset}, and others. Despite recent advancements in 3D generative models, they still face significant challenges in meeting the requirements of real-world applications.

One major limitation of existing models is their inability to generate high-resolution 3D shapes with both quality and diversity. The resulting 3D shapes often suffer from low-level artifacts, lack of fine-grained textures, and insufficient details, which greatly impact their visual fidelity and realism. Additionally, these models exhibit a lack of diversity, often generating only a limited number of variations of similar shapes. This lack of diversity restricts their usability in applications that demand a broad range of high-quality 3D objects, such as gaming, virtual reality, and product design.

In our work, we draw inspiration from the remarkable achievements of large language models (LLMs) \cite{thoppilan2022lamda,ouyang2022training,gpt1,gpt2,gpt3,chowdhery2022palm,zhang2022opt,gpt4,llama} and extend their success to the field of 3D generation. Specifically, we make significant contributions by adapting the Auto-Regressive model introduced in \cite{iccv} and scaling it up to an unprecedented magnitude, named \emph{Argus-3D}, comprising a remarkable 3.6 billion trainable parameters. This groundbreaking endeavor establishes our model as the largest 3D shape generation model known to date.

Our approach revolves around the acquisition of latent representations for 3D shapes, which we accomplish by training our model to learn tri-plane features, effectively reducing computational complexity. Additionally, we introduce a discrete codebook to effectively quantize these representations. Moreover, we employ transformers to predict the quantized representations, leveraging conditions to enable the generation of 3D shapes with multi-modal capabilities.

To facilitate the training of our expansive model, we harness the potential of an ensemble of publicly-available 3D datasets. Our training dataset comprises an impressive collection of nearly 800,000 objects sourced from prominent repositories such as ModelNet40 \cite{modelnet}, ShapeNet \cite{shapenet}, Pix3D \cite{pix3d}, 3D-Future \cite{3dfuture}, and Objaverse \cite{objaverse}.
Our experiments substantiate the efficacy of our approach, as our large-scale 3D generation model significantly enhances the visual quality of the generated 3D shapes. 

Overall, our contributions lie in scaling 3D generation to unprecedented dimensions, developing a remarkable model with a vast number of parameters, exploring novel approaches for latent representation learning, introducing a discrete codebook for effective quantization, and leveraging transformers for multi-modal conditional generation. 
Through these advancements, we pave the way for substantial advancements in the field of 3D shape generation, enabling the generation of high-quality, diverse, and visually impressive 3D shapes.

\section{Related work}
\label{section2}

\textbf{3D Generative Models.}
Recently, there has been significant exploration of 3D shape generative models across various data formats, including polygen meshes\cite{zhang2021sketch2model,iccv,tan2018variational}, point clouds\cite{pointnet,wu2019pointconv,achlioptas2018learning,luo2021diffusion}, voxel grids\cite{choy20163d,voxnet,3dgan,lin2023infinicity,xie2018learning}, and implicit representations such as signed distance functions (SDFs)\cite{chen2019learning,cheng2022cross,autosdf,metasdf,deepsdf,wang2023alto}, among others. Each data format comes with its own advantages and disadvantages, making them suitable for different practical tasks.
(1) Voxel representation inherits the ease of processing through 3D convolutions, similar to 2D images. However, it often suffers from low resolution due to its cubic space occupancy, with existing 3D generative models being limited to resolutions of 64$^{3}$ or lower.
(2) Point clouds, extracted from shape surfaces, do not suffer from resolution restrictions and can handle a larger number of points. However, this representation lacks the ability to preserve 3D topological information and struggles with reconstructing detailed model surfaces.
(3) Mesh-based methods represent 3D objects as a collection of vertices, edges, and faces, forming a polygonal mesh structure. This format is widely utilized due to its ability to represent complex geometry with relatively low storage requirements. However, memory consumption increases significantly as the number of polygons and vertices grows.
(4) Implicit representations excel in representing high-resolution shapes with arbitrary topology. They can predict the signed distance function and recover shape surfaces through techniques like Marching Cubes.
Despite the advancements in these different approaches, each has its own strengths and limitations. The choice of representation depends on the specific requirements of the application at hand.


\textbf{Autoregressive Models.}
Autoregressive models are powerful probabilistic generative models that capture the distribution of probability density by factorizing the joint distribution into a series of conditional distributions using the probability chain rule. Unlike GANs, which lack a tractable probability density and often suffer from generation collisions, autoregressive models exhibit stability during training and have demonstrated their effectiveness in generating natural language, audio, and images\cite{van2016conditional,vqvae2,vqgan}.
However, in the context of 3D shape generation, most autoregressive models face challenges in producing high-quality shapes due to the lack of efficient representation. Recently, DDPMs\cite{ldm,ddpm,ddim,dhariwal2021diffusion} have gained considerable attention for their stability and diverse image generation capabilities. DDPMs learn the data distribution from a Gaussian distribution through a gradual denoising process, often based on the U-Net\cite{unet} architecture. However, this approach limits the resolution and extends the training cycles, thus hindering its performance in 3D generation.
In contrast, transformer-based models offer a high degree of scalability, which has been shown to enhance performance across various downstream tasks by scaling up the model size\cite{wei2022emergent,gpt4,zhai2022scaling,dehghani2023scaling}. We posit that this scalability can also be advantageous in the domain of 3D shape generation, enabling more effective and efficient modeling of complex shapes.

\begin{figure}
    \centering
    \includegraphics[width=\linewidth]{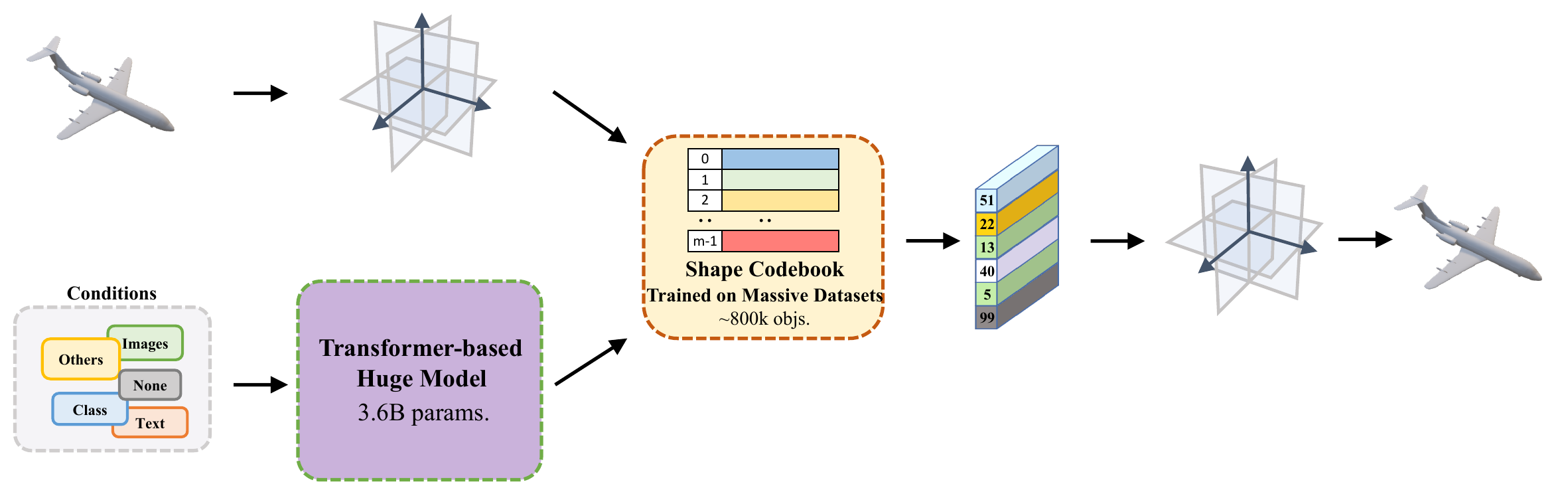}
    \caption{Overview of our model. We derive the shape features from arbitrary 3D shape via encoding points feature into the three normalized orthogonal planes, and project them into a latent vector.  Next, a learnable codebook quantize it for discrete representation. These discrete representation assist the transformer in the second stage to learn the joint distribution corresponding to a large number of shape features encoded in the codebook. Futhermore, by concatenating various conditions such as images or text, transformer is capable of generate discrete representation for 3D shapes. We develop a remarkable 3D shape generalization model with 3.6 billion trainable parameters.}
    \label{fig:main}
\end{figure}

\section{Methodology}
\label{section3}
We adapt the Auto-Regressive model in \cite{iccv} and scale up the learnable parameters. Our model is illustrated in Figure \ref{fig:main}. We train the large model on a collection of 3D datasets. 
Training our 3D shape generation model consists of two stage. It first encodes the input into discrete representation and learns to decode the indices sequence of quantizer. Then, we train a transformer to generate the indices auto-regressively. With the powerful generative capabilities of transformer, our model is able to generate high-quality and diverse 3D shapes.

\subsection{Learning Discrete Representation}

Firstly, we sample point clouds $ \mathcal{P} \in \mathbb{R}$$^{n\times3}$ from mesh as input, and utilize a PointNet\cite{pointnet} to obtain point features with dimension $\mathbb{R}^{n\times32}$. These points are mapped to three axis-aligned orthogonal planes to obtain planar features $\mathbf{f}^v \in  \mathbb{R}^{r\times r\times c\times3}$, where $r$ denote the resolution of feature plane amd $c$ is the dimension. These features are then concatenated and projected to a serialized vector $ \mathbf z_{\mathbf v} \in \mathbb{R}^{(\overline{r}\times\overline{r}) \times d}$ using positional information, which allows us to maintain the order of the feature representations, where $\overline{r} = \frac{r}{8}$ . Later, a learned discrete codebook $\mathbb{Z} = \{ \mathbf z_k\}_{k=1}^K$is used to quantize $\mathcal{Q} (\cdot)$:
\begin{equation}
\label{eq1}
	\mathbf z_{\mathbf q} = \mathcal{Q}(\mathbf z):= arg \min_{\mathbf z_k\in\mathbb{Z}}||\mathbf z_i - \mathbf z_k||
\end{equation}
where $\mathbf z_i \subset \mathbb{R}^d$.
We decode the tri-plane features from the quantized vector using a 2D U-Net. Last, we reconstruct voxel from plane feature and extracted the output shape with Marching Cubes\cite{marching}.

By mapping 3D shape to three orthogonal planes, we can significantly reduce the computational complexity from being proportional to the cube resolution to being proportional to the square resolution, i.e. from $\mathcal{O}(r^3)$ to $\mathcal{O}(r^2)$, while still retaining the important spatial dependencies between points. 
To ensure that the representations retain the spatial dependencies between different points, we use convolutional neural networks (CNNs) to extract features from the planar representations, which are then concatenated and projected to a serialized vector with positional information to maintain the order of the feature representations.
Different from \cite{iccv}, our model does not include the two fully connected layers between the quantized vectors and the encoder and decoder. We remove the two modules because the large model can learn to encode/decode the quantized vectors directly. 

\paragraph{Optimization Objective.}
We optimize the parameters by minimizing the reconstruction error. We apply binary cross-entropy loss between the reconstruction occupancy values $y_o$ and the input values $y_i$:
\begin{equation}
\label{eq2}
	\mathcal{L}_{occ} = -(y_o\cdot log(y_i) + (1 - y_i)\cdot log(1-y_o))
\end{equation}
The codebook loss, employed to minimize the discrepancy between the quantized representation and the feature vectors, is defined as follows:
 \begin{equation}
\label{eq3}
	\mathcal{L}_{quant} = ||{sg}[\mathbf z_{\mathbf v}] - \mathbf z_{\mathbf q}||_2^2 + \beta ||sg[\mathbf z_{\mathbf q}] - \mathbf z_{\mathbf v}||_2^2
\end{equation}
Here, $sg[\cdot]$ denotes the stop-gratient operation, and $||sg[\mathbf z_{\mathbf v}] - \mathbf z_{\mathbf v}||_2^2$ is the \emph{commitment loss} with weighting factor $\beta$, we set $\beta = 0.4 $ by default. The overall reconstruction loss for the first stage is 
\begin{equation}
\mathcal{L}_{rec} = \mathcal{L}_{occ} + \mathcal{L}_{quant}
\end{equation}

\subsection{Learning Shape Generation}
We use a vanilla transformer to learn generate sequence indices of entries in the codebook. This indices represents discrete representation learning in first stage. Each input 3D shape has a tractable order and could be generate as a sequence indices.
A learnable start-of-sequence token is used to predict the first index of the indices order. 
We feed discretized indices of latent vector $\mathbf z = \{\mathbf z_1, \mathbf z_2, \cdot\cdot\cdot, \mathbf z_m\}$ into transformer to learn retrieve the discrete indices. By learning the distribution of previous indices $p(\mathbf z_i | \mathbf z_{<i})$, the model can predict the next index with joint distribution:
\begin{equation}
    p(\mathbf z) = \prod_{i=1}^m p(\mathbf z_i | \mathbf z_{<i})
\end{equation}

\paragraph{Joint Representation Learning.}
By embedding condition information to a vector $\mathbf{c}$ and learn the joint distribution with transformer, our model can generation 3D shape in condition. In lieu of intricate module design or training strategies, we adopt a simpler approach by learning the joint distribution with given conditions $\mathbf c$, achieved by appending it to $\mathbf z$, As follow,
\begin{equation}
    p(\mathbf z) = \prod_{i=1}^m p(\mathbf z_i | \mathbf c, \mathbf z_{<i})
\end{equation}
where $\mathbf c$ denotes a feature vector extracted from arbitrary forms of conditions \textit{i.e.} one-dimensional text, audio, two-dimensional images, or three-dimensional point clouds \textit{etc.}

\paragraph{Optimization Objective.}
To ensure that the generated data closely approximates the real distribution, we minimize the discrepancy between the generated samples and the real data distribution $p(\mathbf x)$:
\begin{equation}
    \mathcal{L}_{nll} = \mathbb{E}_{\mathbf x\sim p(\mathbf x)}[- \log  p(\mathbf z)]
\end{equation}


\subsection{Scaling Up Model}


Motivated by the remarkable progress observed in large language models, we embark on a transformative journey to scale up the 3D generation model, unearthing its untapped potential for substantial performance improvements. In our quest, we present noteworthy contributions by doubling the codebook number and dimension, elevating them to unprecedented heights at 8192 and 512, respectively, surpassing previous pioneering works. Drawing inspiration from the groundbreaking GPT3\cite{gpt3}, we meticulously set the transformer layer, dimension, and head to 32, 3072, and 24, respectively. This deliberate orchestration culminates in the birth of an awe-inspiring 3D generation model, proudly boasting an astonishing 3.6 billion parameters. We denote it as \emph{Ours-Huge}. We also derive a smaller version with 1.2 billion parameters for fast experiments, which is named \emph{Ours-Large}. The only difference is that we set transformer layer, dimension, and head to be 24, 2048 and 16.

To the best of our knowledge, we highlight that our proposed model stands as the largest of its kind in the realm of 3D shape generation, eclipsing the parameter counts of renowned works such as SDFusion~\cite{cheng2023sdfusion} and Shap-E~\cite{jun2023shap}, which consist of approximately 1 billion parameters. This monumental achievement in scaling up the model represents a significant milestone, propelling the field of 3D shape generation towards unparalleled advancements.





\subsection{Scaling Up Data}

In our quest to build a robust and diverse training dataset, we draw inspiration from the remarkable progress observed in the realm of 2D vision, where the effectiveness of large models fueled by ample training data has been extensively demonstrated \cite{dosovitskiy2021image, liu2021swin, fang2022eva}. However, we acknowledge the inherent challenges associated with constructing large-scale 3D shape datasets, as the process of labeling 3D objects entails considerably higher costs and complexities compared to their 2D counterparts.

In this paper, we proudly present our significant contributions, which revolve around the meticulous curation of a diverse and comprehensive collection of publicly-available 3D datasets. These datasets serve as invaluable resources for training and evaluation, enabling us to push the boundaries of 3D shape generation. Our curated dataset comprises:

\textbf{ModelNet40}\cite{modelnet}: This dataset stands as a valuable resource, containing an impressive collection of over 12,300 CAD models spanning 40 object categories. For our research objectives, we strategically select 9,843 models from this dataset to serve as the foundation for training our model. An additional 2,468 models are reserved for rigorous testing, enabling us to comprehensively evaluate the performance and generalization capabilities of our approach.

\textbf{ShapeNet}\cite{shapenet}: Recognized for its comprehensiveness and scope, ShapeNet provides an extensive and diverse collection of 3D models. Encompassing 55 common object categories and boasting over 51,300 unique 3D models, this dataset offers rich and varied representations that greatly enhance the training process of our model. The inclusion of ShapeNet in our dataset selection bolsters the ability of our model to capture a wide range of object variations and complexities.

\textbf{Pix3D}\cite{pix3d}: With its large-scale nature, Pix3D emerges as a valuable resource for our research. Comprising 10,069 images and 395 shapes, this dataset exhibits significant variations, allowing our model to learn from a diverse range of object appearances and characteristics. By incorporating Pix3D into our training dataset, we equip our model with the capability to generate visually impressive and realistic 3D shapes.

\textbf{3D-Future}\cite{3dfuture}: As a specialized furniture dataset, 3D-Future provides a unique and detailed collection of 16,563 distinct 3D instances. By including this dataset in our training pipeline, we ensure that our model gains insights into the intricacies and nuances of furniture designs. This enables our model to generate high-quality and realistic 3D shapes in the furniture domain, facilitating its application in areas such as interior design and virtual staging.

\textbf{Objaverse}\cite{objaverse}: A recent addition to the realm of 3D shape datasets, Objaverse captivates us with its vastness and richness. Sourced from Sketchfab, this immensely large dataset comprises over 800,000(and growing) 3D models, offering a wide array of object categories and variations. This extensive dataset providing our model with a wealth of diverse and representative examples for learning and generating high-quality 3D shapes.

Through meticulous curation and the thoughtful incorporation of these prominent datasets, we ensure that our training process encapsulates a diverse and comprehensive spectrum of 3D object variations, laying the groundwork for remarkable advancements in the field of 3D shape generation.

In our endeavor, we amass an impressive collection comprising approximately 900,000 3D shapes. To prepare the data, we meticulously follow the established methodologies outlined in 3D-R2N2\cite{choy20163d} and C-OccNet\cite{occupancy}, employing techniques such as mesh-fusion\cite{meshfusion} to create depth maps, generating watertight meshes, and sampling points from these meshes. It is worth noting that the generation of intermediate files from a single ShapeNet dataset can exceed 3TB during these intricate steps. For the Pix3D chair category, we leverage the provided train/test splits. Regarding the other datasets, we randomly select the 80\% as training samples, while the remaining 20\% is reserved for testing purposes. Due to the sheer size of the objaverse dataset, we selected only 10\% of the dataset as the dedicated test set. All of these carefully curated datasets play a pivotal role in training our first-stage model, serving as the bedrock of our research endeavor.

\section{Experiments}
\label{section4}


In our rigorous evaluation process, we employ the Intersection over Union (IoU) metric to assess the reconstruction performance of the first stage. Once the IoU surpasses the remarkable threshold of 0.9, a pivotal moment arrives, as we leverage the trained codebook to quantize the input data, ushering forth the generation of corresponding index sequences. These meticulously quantized data then become the bedrock for training the autoregressive model of the second stage, fueling its remarkable capabilities.

To comprehensively gauge the prowess of our model, we embark on an extensive array of four distinctive experiments. Firstly, we delve into the realm of unconditional generation, showcasing our model's remarkable aptitude in crafting intricate shapes across four widely-used categories: planes, chairs, cars, and tables. Emboldened by this achievement, we venture further into the domain of conditional generation, where our model shines even brighter, effortlessly producing an array of faithful and diverse shapes across several intriguing tasks. The culmination of these experiments unveils the true potential of our model, positioning it at the forefront of cutting-edge 3D shape generation research.

\subsection{Unconditional Generation}
\label{uncond}


In the landscape of prior research\cite{iccv,autosdf,ibing20213d}, a prevailing approach involves training unconditional generation models on a per-category basis, subsequently evaluating their performance by generating samples exclusively from that specific category. This conventional framework, however, necessitates the training of separate models for each category, imposing substantial costs when working with large-scale models. Moreover, the relatively limited availability of data per category inherently constrains the potential and efficacy of our expansive models.

\begin{figure}
    \centering
    \includegraphics[width=0.5\linewidth]{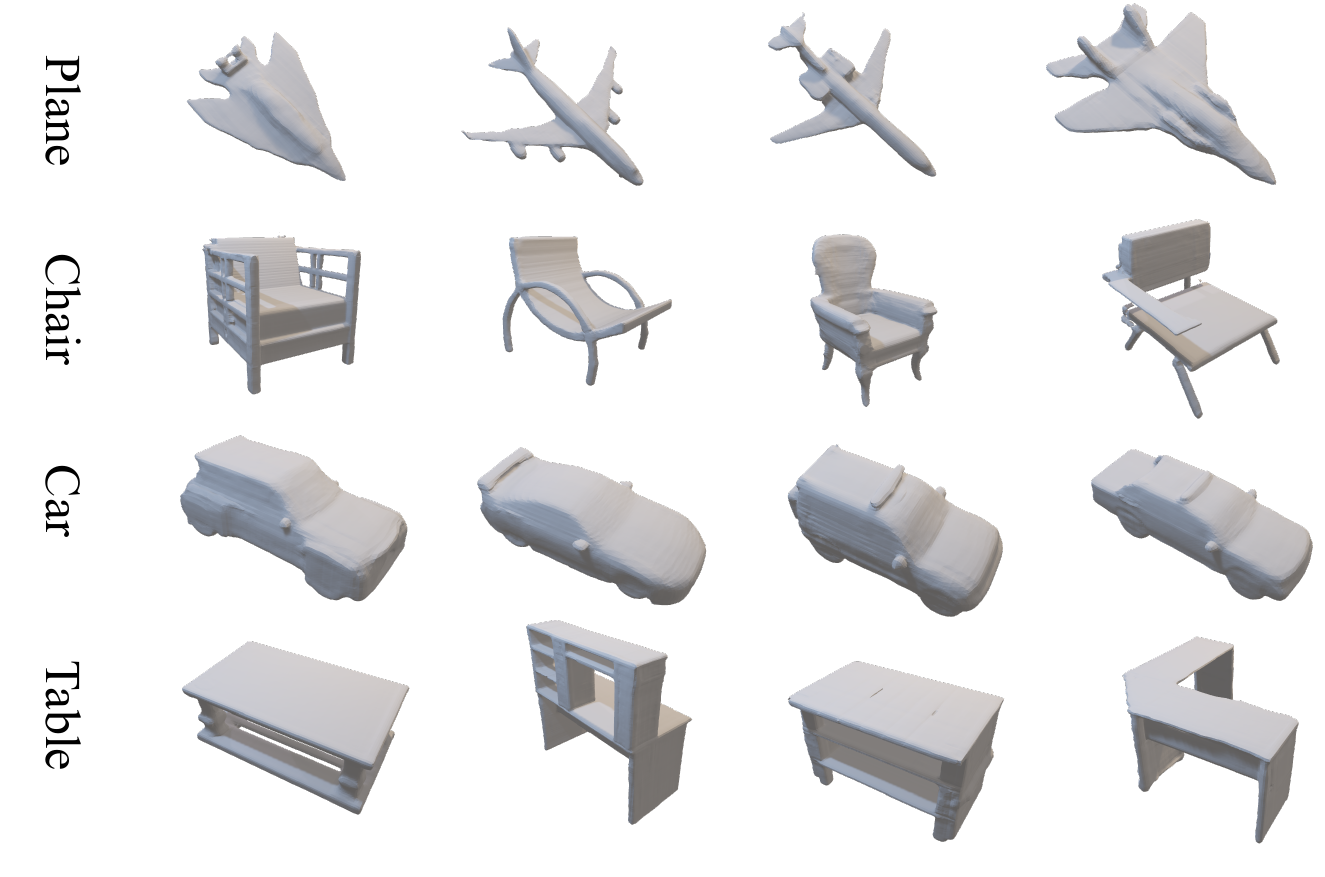}
    \caption{Qualitative results of unconditional generation.}
    \label{fig:uncond}
\end{figure}

In light of these challenges, we adopt a pragmatic strategy, leveraging our insights to train multiple smaller models on commonly encountered categories. This judicious approach enables us to comprehensively assess the generative prowess of our models while circumventing the limitations imposed by category-specific training. The captivating results of our efforts are eloquently showcased in the accompanying Figure ~\ref{fig:uncond}, where the rich diversity and remarkable fidelity of the generated samples beautifully exemplify the transformative capabilities of our approach. 


\begin{figure}
    \centering
    \includegraphics[width=0.5\linewidth]{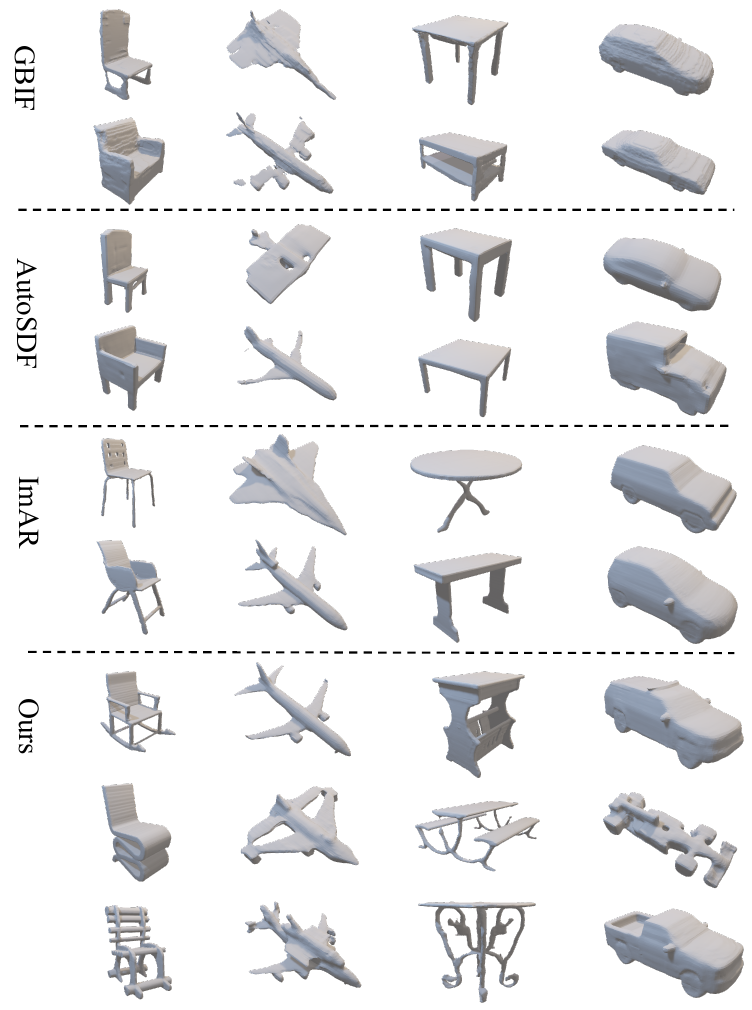}
    \caption{Qualitative results of class-guide generation.}
    \label{fig:class}
\end{figure}

\subsection{Class-guide Generation}



In our pursuit of enhancing the generative capabilities of our model, we undertake a comprehensive training approach that incorporates class conditioning across all 55 diverse categories from the ShapeNet dataset. This endeavor necessitates the availability of category labels to guide the shape generation process, facilitating the synthesis of shapes that align with the specific category characteristics.

Building upon the methodologies established by previous works~\cite{ibing20213d,chen2019learning}, we rigorously evaluate the performance of our model using a range of insightful metrics. The 1-Nearest Neighbor Accuracy (1-NNA)~\cite{lopez2016revisiting,pointflow} serves as our primary metric for measuring the distribution similarity, employing both the Chamfer distance (CD) and earth mover distance (EMD). Additionally, we utilize Coverage (COV)~\cite{achlioptas2018learning}, Minimum Matching Distance (MMD)~\cite{achlioptas2018learning}, and Edge Count Difference (ECD)~\cite{ibing20213d} to assess the diversity, fidelity, and overall quality of the synthesized shapes.

To benchmark the performance of our model, we compare it against three baseline models, namely GBIF~\cite{ibing20213d}, AutoSDF~\cite{autosdf}, and ImAR~\cite{iccv}. As illustrated in Table \ref{tab:class}, our huge model consistently outperforms the baselines across various metrics, signifying a substantial improvement in overall performance. It is worth noting that our model exhibits slightly lower performance in terms of MMD for the plane class. One possible explanation for this observation is that our model's encoding of 3D shapes to planar features may weaken the spatial constraints on flat objects, resulting in the generation of partially scattered points.

We visualize generated shapes with multiple categories in Figure \ref{fig:class}. 
Our generated shapes are characterized by intricate details and complex structures, which also significantly contribute to the diversity of the results. To fully appreciate the effect, please zoom in and observe the fine-grained intricacies.

                    
\begin{table}                               
\centering  \small                              
\setlength{\tabcolsep}{6pt}   
\begin{tabular}{cccccccccc}                                                 
\toprule                       
&  &\multicolumn{2}{c}{MMD($\times 10^3$)$\downarrow$}&\multicolumn{2}{c}{COV(\%)$\uparrow$}&\multicolumn{2}{c}{1-NNA(\%)$\downarrow$}&\multicolumn{2}{c}{ECD$\downarrow$}
\\  \cmidrule(lr){3-4} \cmidrule(lr){5-6} \cmidrule(lr){7-8} \cmidrule(lr){9-10}
Category   &Model  &CD  &EMD   &CD   &EMD   &CD   &EMD 	&CD   &EMD 			
\\  \midrule                                       
\multirow{5}{*}{Plane}      & GBIF \cite{ibing20213d} & 1.4394          & 1.1348          & 42.33          & 45.30           & 89.23          & 79.83          & 1563         & 1355         \\
 & AutoSDF \cite{autosdf}  & 4.8047          & 2.4814          & 26.57          & 31.32          & 80.66          & 79.30           & 4750         & 5101         \\
 & 3DILG \cite{zhang20223dilg}   & 1.3995          & 0.9774          & 39.11          & 47.52          & 73.27          & 59.16           & 197         & 168         \\
      & ImAM \cite{iccv}        & \textbf{1.0577} & \textbf{0.8587} & 49.01          & 50.99          & 63.49          & 55.32          & 43           & 92           \\ \cmidrule{2-10} 
      & \textit{Ours}           & 1.0998           & 0.8680          & \textbf{50.25} & \textbf{56.68} & \textbf{58.66} & \textbf{48.89}  & \textbf{2}  & \textbf{58}  \\ \midrule
\multirow{5}{*}{Chair}      & GBIF \cite{ibing20213d} & 4.2520           & 2.3369          & 49.63          & 52.73          & 68.46          & 62.85          & 184          & 290          \\
 & AutoSDF \cite{autosdf}  & 5.5206          & 2.7157          & 34.71          & 37.67          & 81.39          & 76.29          & 3166         & 2991         \\
 & 3DILG \cite{zhang20223dilg}   & 5.1374          & 2.5965          &  41.72          & 42.31          & 65.98          & 64.05          & 323         & 351         \\
      & ImAM \cite{iccv}        & 4.1144          & 2.2620           & 49.63          & 51.85         & 59.90           & 56.94          & 79           & 183          \\ \cmidrule{2-10} 
      & \textit{Ours}           & \textbf{3.7802} & \textbf{2.1334} & \textbf{51.55} & \textbf{53.62}  & \textbf{54.51} & \textbf{52.36} & \textbf{6}   & \textbf{95}  \\ \midrule
 \multirow{5}{*}{Table}     & GBIF \cite{ibing20213d} & 3.4813          & 2.1010           & 50.89          & 53.86          & 66.67          & 62.99          & 98           & 457          \\
 & AutoSDF \cite{autosdf}  & 4.8047          & 2.4814          & 26.57          & 31.32          & 80.66          & 79.30           & 4750         & 5101         \\
  & 3DILG \cite{zhang20223dilg}  & 5.745          & 2.6694          & 28.98          & 32.78          & 75.89          & 74.47           & 1709         & 1635         \\
      & ImAM \cite{iccv}        & 3.3972          & 1.9795          & 48.64          & 53.86          & 56.17          & 53.68          & 52           & 114          \\ \cmidrule{2-10} 
      & \textit{Ours}           & \textbf{3.0622} & \textbf{1.8813} & \textbf{54.45}  & \textbf{54.21} & \textbf{50.83} & \textbf{50.18}  & \textbf{1}   & \textbf{87}  \\ \midrule
 \multirow{5}{*}{Car}     & GBIF \cite{ibing20213d} & 1.2853          & 0.9062          & 33.71          & 40.00             & 90.00             & 79.14          & 2453         & 1471         \\
  & AutoSDF \cite{autosdf}  & 1.2695          & 0.8739          & 34.86          & 41.71          & 88.29          & 80.71          & 2312         & 1299         \\
    & 3DILG \cite{zhang20223dilg} & 1.2914          & 1.0398         & 26.86          & 31.43          & 81.43          & 75.71          & 847         & 741         \\
      & ImAM \cite{iccv}        & 1.1929          & 0.8704          & 36.45          & 37.38          & 73.77          & 72.43          & 1823         & 2231         \\ \cmidrule{2-10} 
      & \textit{Ours}           & \textbf{1.0901} & \textbf{0.8104} & \textbf{40.57} & \textbf{42.00} & \textbf{72.14}    & \textbf{65.14} & \textbf{472} & \textbf{430} \\  \midrule 
\multirow{5}{*}{Avg.}        & GBIF \cite{ibing20213d} & 2.6145          & 1.6197          & 44.14          & 47.97          & 78.59          & 71.20          & 1075         & 894          \\
  & AutoSDF \cite{autosdf}  & 3.3977          & 1.9046          & 28.62          & 32.94          & 84.93          & 81.48          & 3669         & 3535         \\
   & 3DILG\cite{zhang20223dilg}   & 3.3933          & 1.8208          & 34.18          & 38.51          & 74.14          & 68.35          & 769         & 724         \\
        & ImAM \cite{iccv}        & 2.4406          & 1.4927          & 45.93          & 48.52          & 63.33          & 59.59          & 500          & 656          \\ \cmidrule{2-10} 
        & \textit{Ours}           & \textbf{2.2581} & \textbf{1.4232} & \textbf{49.21} & \textbf{51.63} & \textbf{59.04} & \textbf{54.15} & \textbf{120} & \textbf{168} \\
\bottomrule                                                                          
\end{tabular}                                         
\caption{ Results of class-guide generation. Models are trained on 55 categories of ShapeNet and generate shapes with categories label.}
\label{tab:class}                      
\end{table}

\subsection{Image-guide Generation}
\label{img-guide}




Undertaking the task of image-guided generation presents a greater challenge, necessitating the utilization of a pretrained CLIP model to extract 2D feature vectors from images, which serve as conditions for our model.

To generate quantized sequences, we make use of all the datasets from the training set of the first stage, excluding ModelNet. These quantized sequences are then paired with randomly rendered multi-view images, resulting in single-view image-shape pairs that serve as training data.

Following the established approach in previous works, we evaluate our models by randomly sampling 50 single-view images from the test split of 13 categories. For each image, we generate 5 shapes for evaluation purposes. To assess the diversity, we employ the Total Mutual Difference (TMD) metric, while the faithfulness of completed shapes is measured using MMD and Fréchet Point Cloud distance (FPD)\cite{shu20193d}.

\begin{figure}
   \centering
   \includegraphics[width=0.7\linewidth]{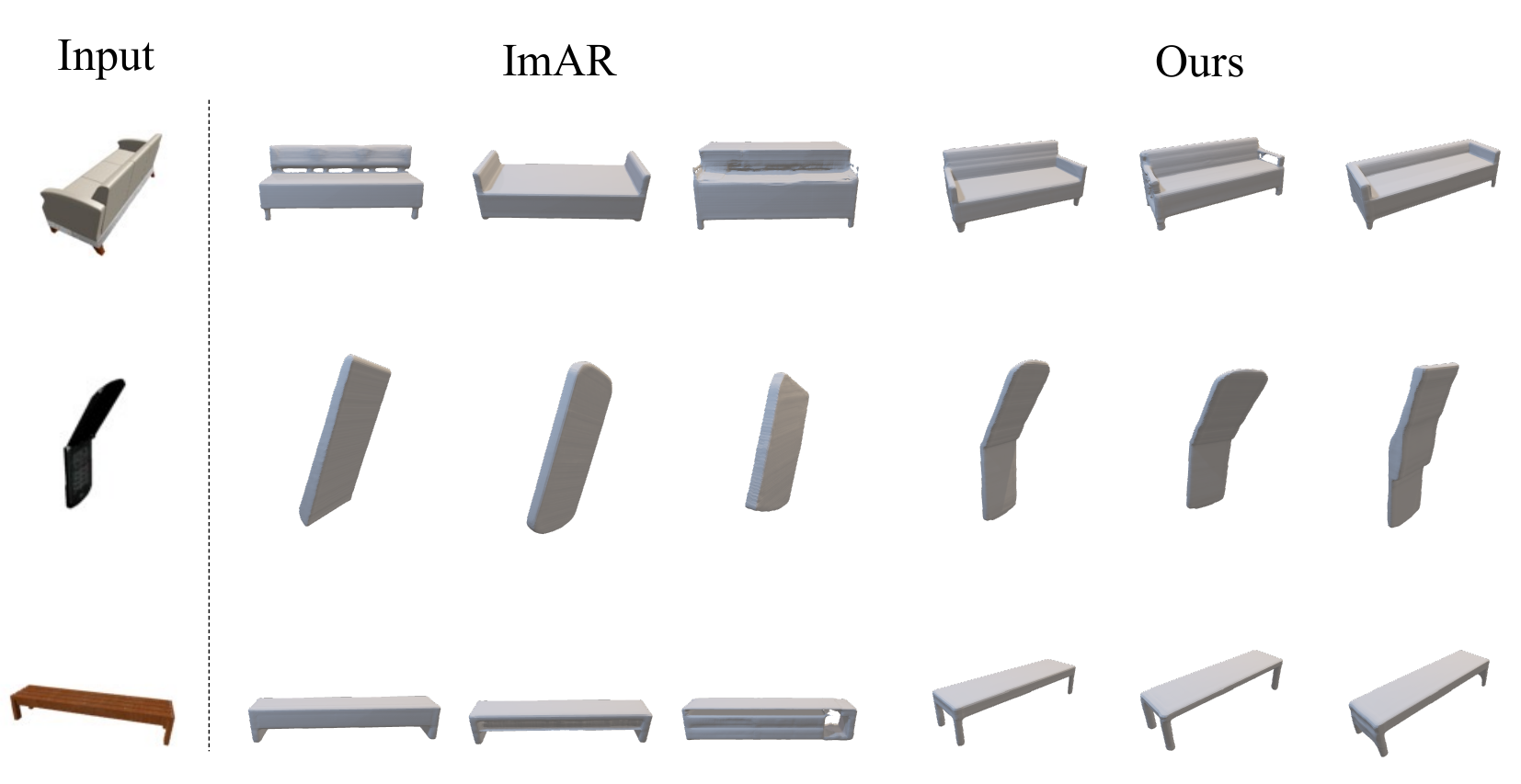}
   \caption{Visualizations of image-guide shape generation.}
   \label{fig:img}
\end{figure}

Table \ref{tab:image_generation} show our large model achieves state-of-the-art performance on image-guide shape generation.  It is worth noting that our image-guided model not only achieves higher fidelity than previous state-of-the-art methods but also significantly boosts diversity in performance.
Qualitative results of our image-guided generation approach are presented in Figure \ref{fig:img}.

\begin{figure}
   \centering
   \includegraphics[width=0.5\linewidth]{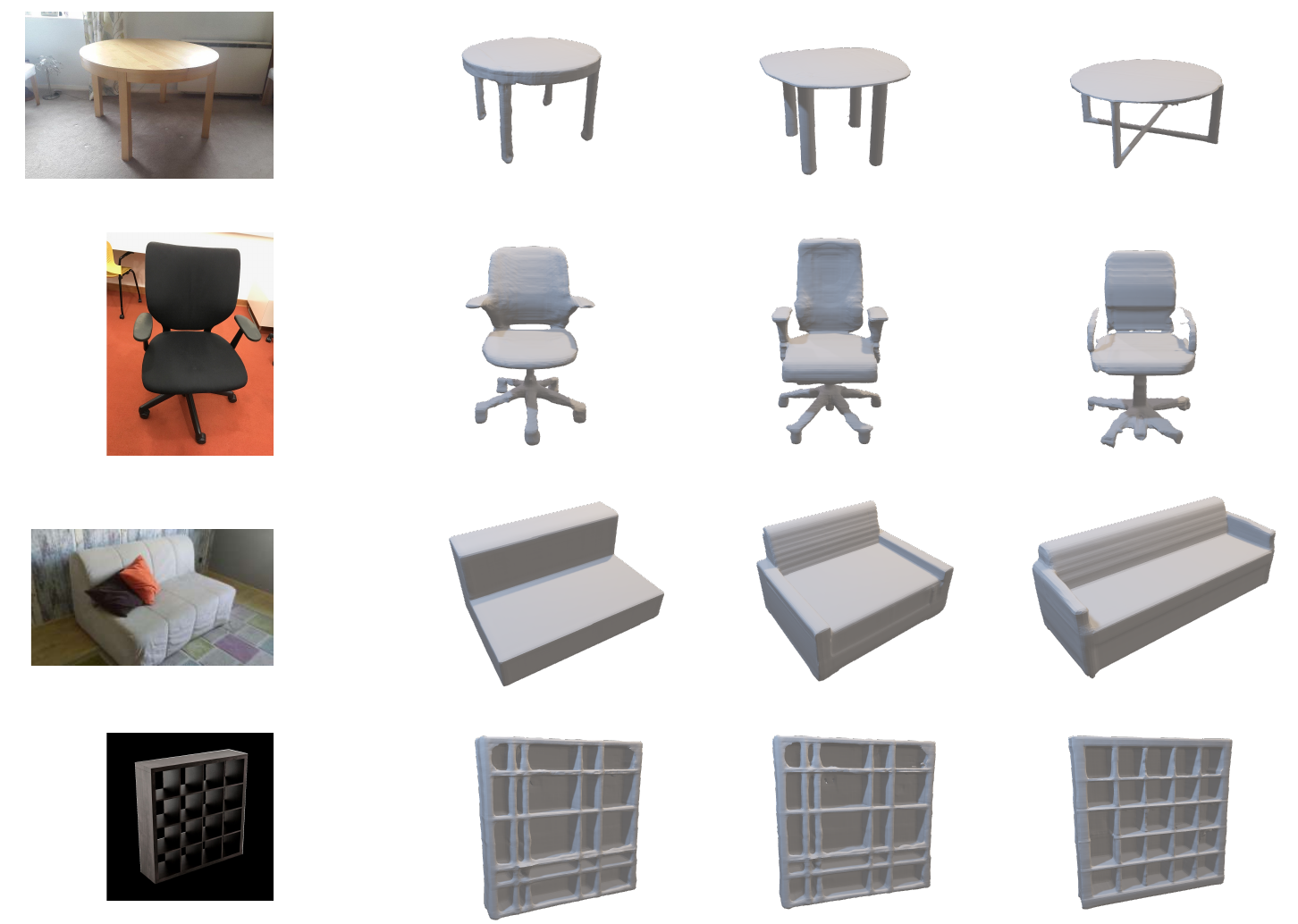}
   \caption{Results of real-world image-guide generation.}
   \label{fig:real}
\end{figure}

Additionally, we explore the generalizability of our model on real-world images. By conditioning our model on Pix3D images, we demonstrate its ability to accurately capture the primary attributes of objects depicted in the images, generating highly quality associated 3D shapes. The remarkable results of this investigation are showcased in the accompanying Figure\ref{fig:real}. 



\begin{table}
\centering
\footnotesize
\setlength{\tabcolsep}{2.5mm}{
\begin{tabular}{cccc} 
\hline 
\multicolumn{1}{c}{\textsc{Method}} & TMD ($\times 10^2 $) $\uparrow$ & MMD ($\times 10^3$) $\downarrow$ & FPD $\downarrow$ \tabularnewline 
\hline 
AutoSDF \cite{autosdf} & 2.657 & 2.137 & 15.262 \tabularnewline 
CLIP-Forge \cite{sanghi2022clip} & 2.858 & 1.926 & 8.094   \tabularnewline 
ImAM \cite{iccv} & 4.274 & 1.590 & 1.680  \tabularnewline 
\textit{Ours} & \textbf{5.136} & \textbf{1.338} & \textbf{0.774}  \tabularnewline 
\hline
\end{tabular}}
\caption{Quantitative results of image-guide generation. \label{tab:image_generation}}
\end{table}

\subsection{Text-guide Generation}

By harnessing the remarkable bridging capabilities of the CLIP model between text and image modalities, our model transcends the boundaries of traditional input sources and enables the generation of 3D shapes solely from zero-shot textual input. During the training phase, we adhered to the methodology outlined in \ref{img-guide}, utilizing paired image-shape inputs. In the inference stage, we took a groundbreaking approach by substituting the image features with text features extracted from CLIP.

\begin{figure}
\begin{minipage}[t]{0.49\textwidth}
    \centering
    \includegraphics[width=1.0\linewidth]{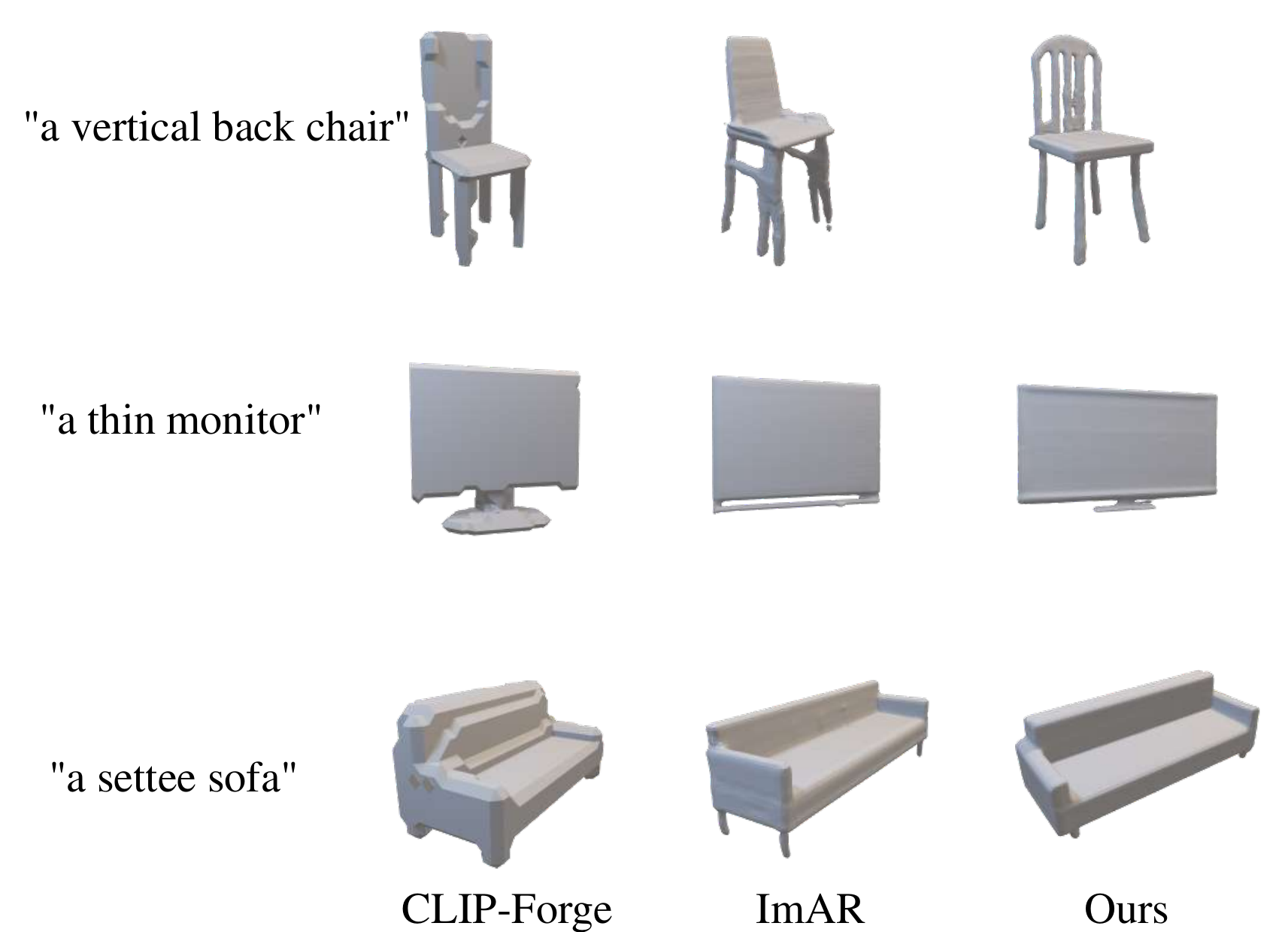}
    \caption{Qualitative comparisons with the previous zero-shot methods.}
    \label{fig:text1}
\end{minipage}
\quad
\begin{minipage}[t]{0.49\textwidth}
    \centering
    \includegraphics[width=1.0\linewidth]{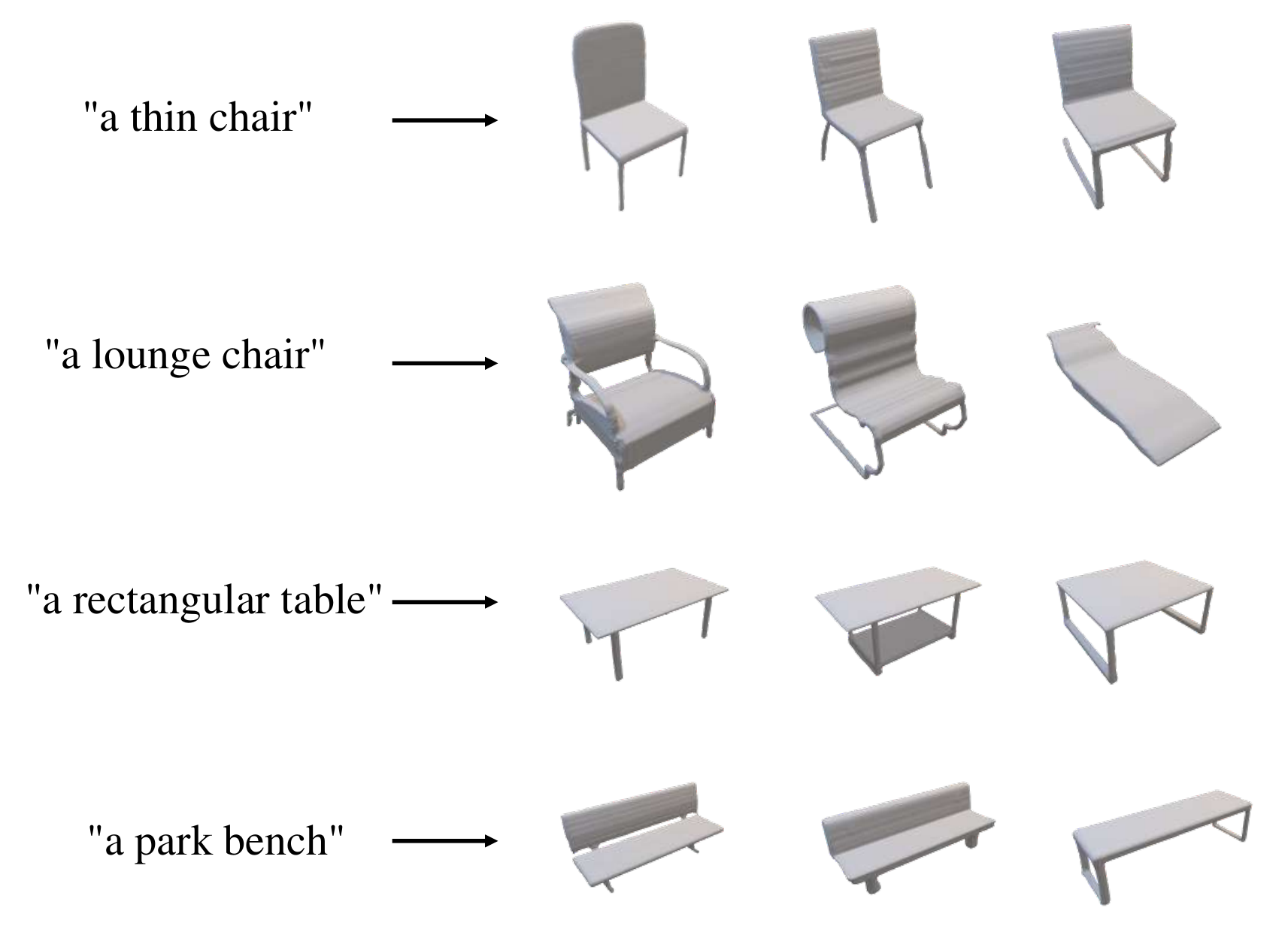}
    \caption{Results of zero-shot text-to-shape generation. }
    \label{fig:text2}
\end{minipage}
\end{figure}

The accompanying figure serves as a visual testament to the impressive generative prowess of our model, showcasing the diverse and captivating 3D shape outputs generated in response to different textual prompts. This groundbreaking advancement exemplifies the power and versatility of our model, as it seamlessly operates across multiple input modalities, opening up exciting new avenues in the realm of 3D shape generation.
As shown in Figure \ref{fig:text1}, our model excels at generating fine-grained shapes, surpassing previous zero-shot approaches. Additionally, we showcase the diversity of our text-guided shape generation through Figure \ref{fig:text2}

\paragraph{Validation of Generative Capacity}
In order to further demonstrate that our huge model has learned the underlying distribution of the 3D shape rather than simply memorizing the training set to reconstruct shape, we selected the generated samples and calculated their nearest neighbors in the training set based on Chamfer Distance. Figure \ref{fig:aa} show our results.




\section{Conclusion}
\label{section5}


In this paper, we explore the limits of 3D shape generation by scaling the adapted Auto-Regressive model to 3.6 billion parameters and train it on the collected dataset with approximately 900,000 objects.
Our larger-scale dataset and increased model parameters enable us to generate a wider variety and complexity of spatial structures, particularly intricate objects like chairs. The comprehensive dataset captures diverse designs and styles, allowing our model to learn underlying patterns effectively.

However, challenges remain, including the need for ample training data and the computational demands of the transformer architecture. To address these, we're exploring domain-specific knowledge integration, more efficient transformer architectures, and novel 3D shape data representations. These efforts aim to enhance our approach's performance and overcome existing challenges.

\medskip

{\small
\bibliographystyle{plain}
\bibliography{references}
}

\newpage
\renewcommand\thesection{\Alph{section}}
\setcounter{section}{0}
\large{\textbf{Appendix}}


\section{Technical Details}
\subsection{Model Architectures}
The architectures of our model, consisting of an encoder, a quantizer and a decoder at the first stage, a transformer at the second stage.
\paragraph{Encoder}
The encoder takes point clouds $ \mathcal{P} \in \mathbb{R}$$^{n\times3}$ as input, while $n$ means the number of points. $n$ should less than 100,000, as we have sampled a total of 100,000 points from the watertight mesh in the data preprocessing stage. Then feeding point clouds into a PointNet with local pooling to result point features. Since the processed points can still be considered to exist in 3D space, in order to reduce computational complexity while retaining spatial position information, we projected these points onto three axis-aligned orthogonal planes as tri-planar features instead of volumetric features. Next, we down sample the feature with convolution layers and output a feature vector.
\paragraph{Quantizer}
We utilize a shape codebook to quantize encoded vector. To avoid information bottlenecks and enable the model to learn better representations, we have set the codebook with 8192 entries, 512 dimensions. The output of the quantizer is a quantized vector composed of 1024 discrete values, while each discrete value corresponds to the entry of the corresponding index in the codebook. 
\paragraph{Decoder}
Firstly, the decoder reshape the quantized vector and a 2D U-Net is used to map discrete values to continuous space. Then we upsample the features back to feature plane with convolution layers. We utilize an implicit function to predict the occupancy probability of each position in the tri-planar features. The final 3D shapes are extracted using the Marching Cubes algorithm, which extracts the isosurfaces from the occupancy grid representation of the generated shapes. This method allows us to obtain a smooth and continuous representation of the shapes, enabling further analysis and visualization.
\paragraph{Transformer}
We use the transformer to model the conditional joint distribution and generate 3D shape in an autoregressive manner. 
For class-guide generation, we extract features from categories use an embedding layer. For image-guide generation, we utilize pre-trained CLIP model to extract image features. To predict the first output index, we use a learnable start-of-sequence token prepended with conditional features. Then model can predict next indices with previous indices regressively.  

\subsection{Training and Testing Procedures}
\paragraph{Training}
In the first stage, we sample point clouds with $n = 30,000$ as input, train the shape codebook with 55 categories on ShapeNet dataset for 400k iterations as checkpoint A. Subsequently, we continued training the models using a larger dataset with over 200,000 object from ShapeNet, ModelNet40, Pix3D, 3D-Future, and Objaverse for 1,100k iterations, make checkpoint B. The learning rate is set as 1e-4, and the batch size is 64. The ckeckpoint A is shared for unconditional generation and class-guide generation task. The checkpoint B is used to train our image-guide model. For the second stage, we adopt different learning rate to train different sizes of models. The learning rates for Small, Large, Huge models are 1e-4, 3e-5, 1e-5, respectively.
\paragraph{Testing}
During the inference process, our trained auto-regressived model predict the discrete index sequence with conditional inputs. Starting from the initial input, the model generates sequence indices one by one. It employs self-attention mechanism to focus on previously generated sequence parts and predicts the next sequence indices. Based on the predicted probability distribution, model kept top-$k$ indices with the highest confidence and sampling to obtain the current index. The process continues to generate the next index until the entire sequence is generated. Finally, we utilize the trained decoder from the first stage to decode this discrete index sequence into a 3D shape.

\section{Ablation Study}
In order to investigate the impact of increasing model parameters on performance, we followed the parameter configuration of GPT-3 and trained three models of different sizes:  Small(100M), Large(1.2B), and Huge(3.6B), both based on first stage shape codebook with 8192 entries, 512 dimensions. Table \ref{tab:para}  illustrates the parameter comparison among our different models.

\begin{table}
\centering
\footnotesize
\setlength{\tabcolsep}{4mm}{
\begin{tabular}{ccccc} 
\toprule 
\multicolumn{1}{c}{\textbf{Size}} & \textbf{Dimensions} & \textbf{Layers} & \textbf{Heads} & \textbf{Params [M]} \tabularnewline 
\midrule
\textit{Small} & 768 & 12 & 12 & 100   \tabularnewline 
\textit{Large} & 2048 & 24 & 16 & 1239  \tabularnewline 
\textbf{\textit{Huge}} & 3072 & 32 & 24 & 3670  \tabularnewline 
\bottomrule
\end{tabular}}
\caption{Model architecture details. \label{tab:para}}
\end{table}

\begin{figure}
    \centering
    \includegraphics[width=0.6\linewidth]{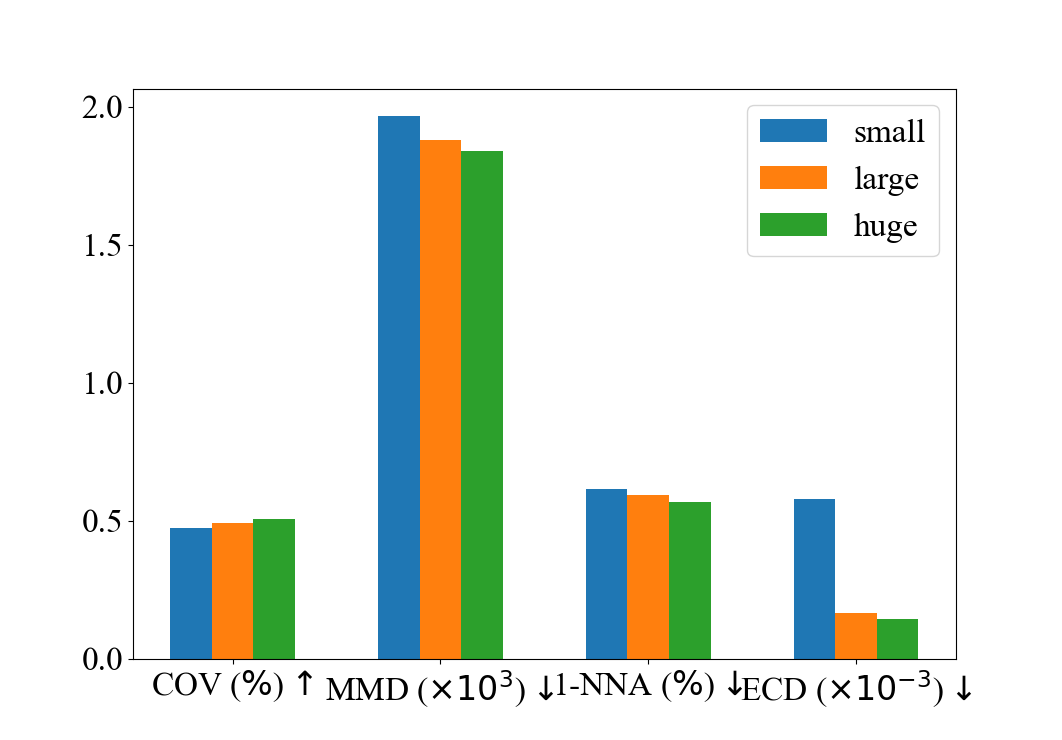}
    \caption{Quantitative result for different models.}
    \label{fig:cmp}
\end{figure}

We evaluated the different models on class-guide generation task. As show in Figure \ref{fig:cmp}, our huge model achieved better performance across multiple metrics. Especially in terms of ECD indicators, larger models have shown significant improvement compared to smaller models.

\begin{figure}
    \centering
    \includegraphics[width=\linewidth]{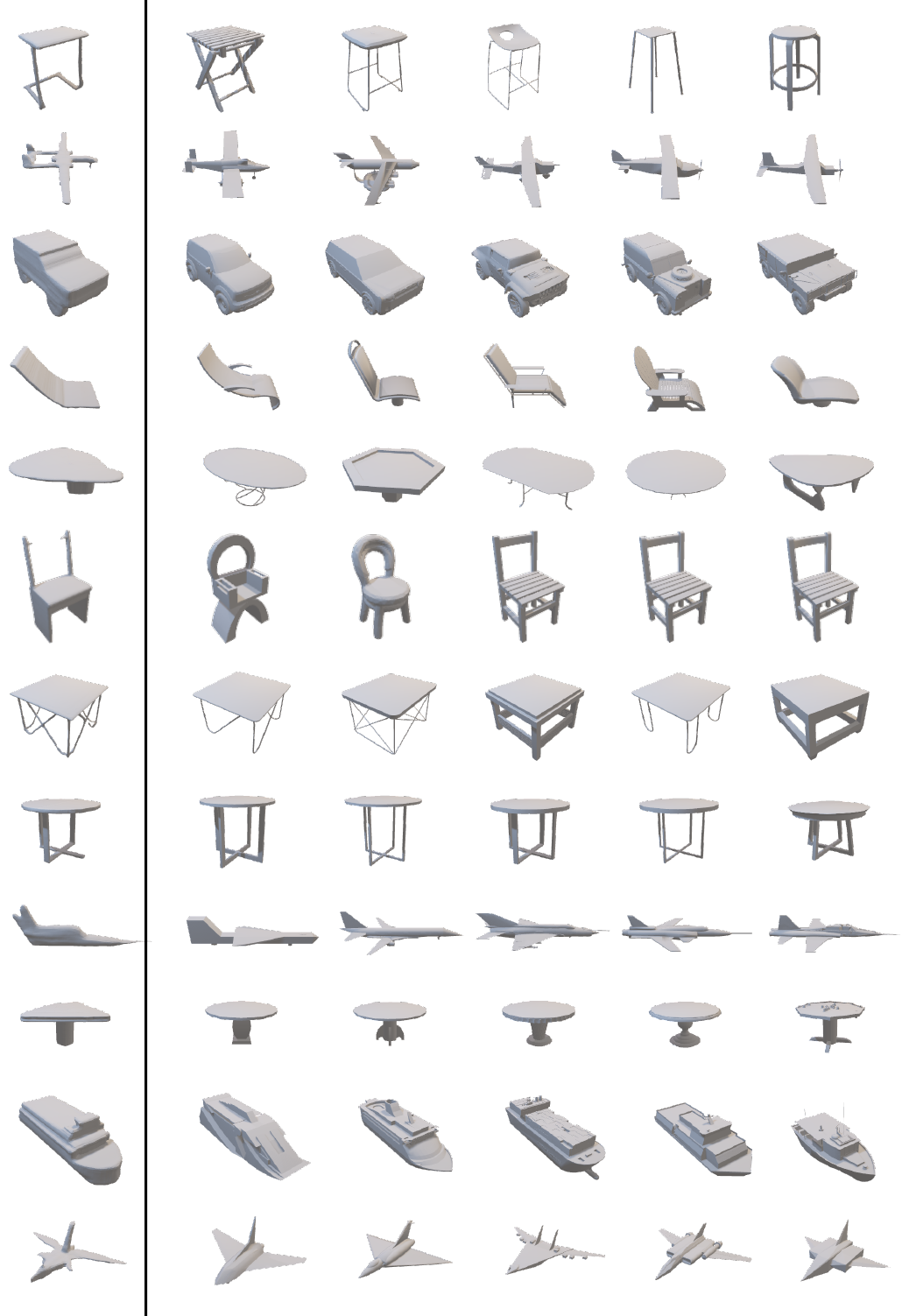}
    \caption{Generated shapes with their most similar shapes in the training set.}
    \label{fig:aa}
\end{figure}

\paragraph{Validation of Generative Capacity}
In order to further demonstrate that our huge model has learned the underlying distribution of the 3D shape rather than simply memorizing the training set to reconstruct shape, we selected the generated samples and calculated their nearest neighbors in the training set based on Chamfer Distance. Figure \ref{fig:aa} show our results.

\section{Qualitative Results}
\subsection{Class-guide Generation}
\begin{figure}
    \centering
    \includegraphics[width=\linewidth]{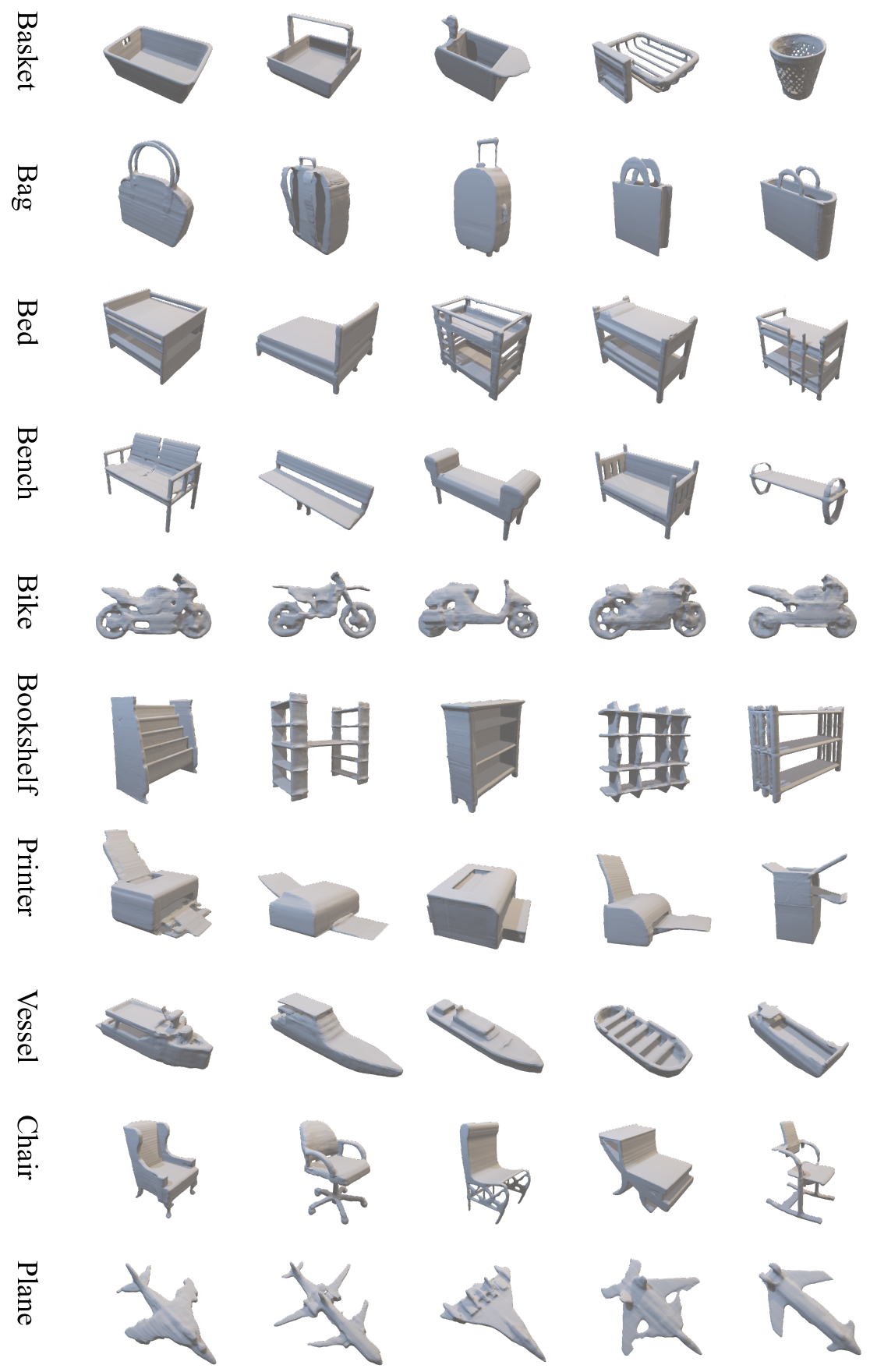}
    \caption{More qualitative results of class-guide generation.}
    \label{fig:my_label}
\end{figure}

We show more visualizations results in Figure \ref{fig:my_label}, where the generation of high-quality and rich diversity shapes across multiple categories demonstrates the powerful generative capacity of our huge model.

\subsection{Image-guide Generation}
\begin{figure}
    \centering
    \includegraphics[width=0.93\linewidth]{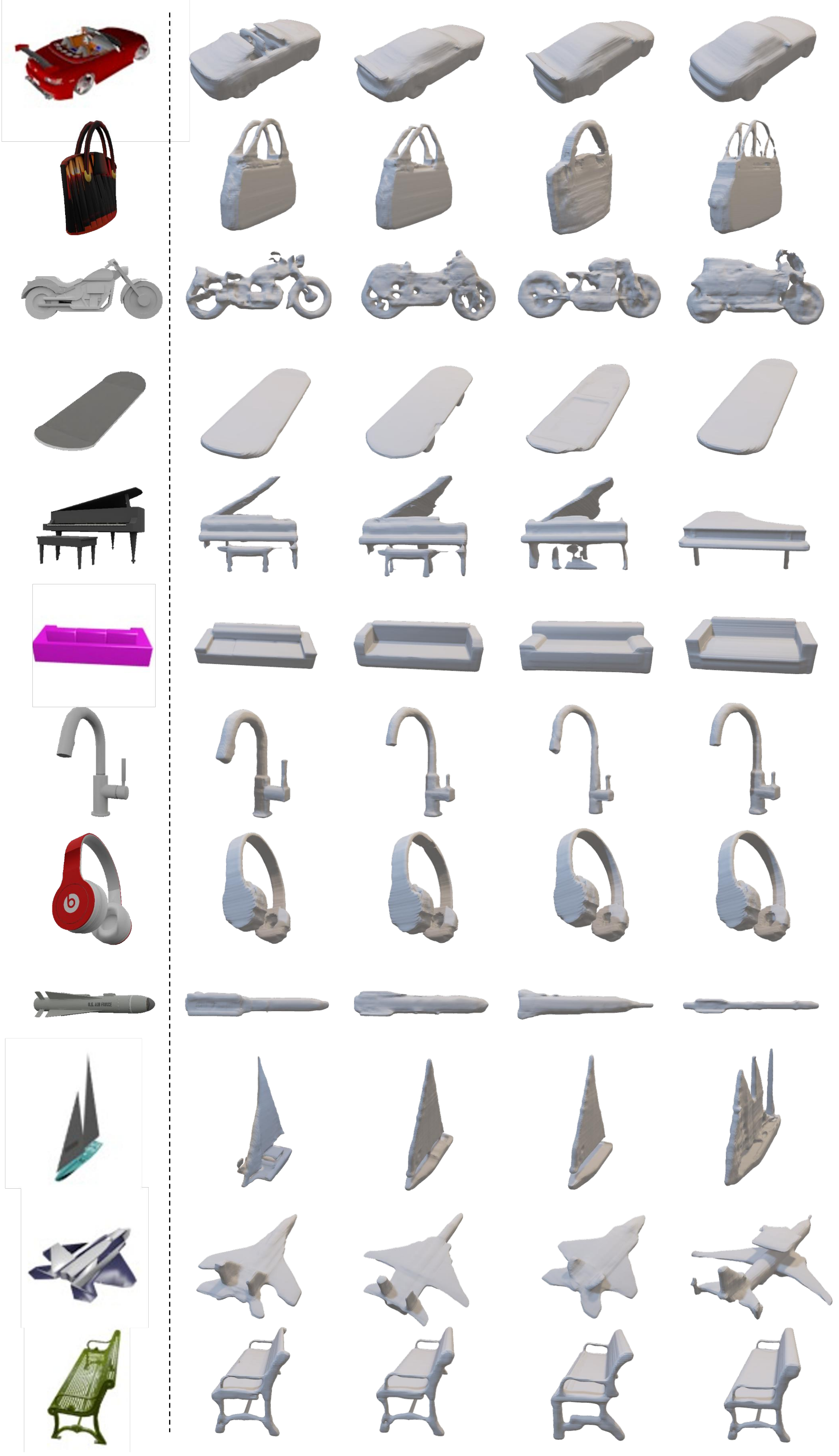}
    \caption{More qualitative results of image-guide generation.}
    \label{fig:img2}
\end{figure}

Figure \ref{fig:img2} present more exemplary results that demonstrate the exceptional performance of our large model in image-guide shape generation.

\section{Broader Impact and Limitation}
\paragraph{Broader Impact}
Synthesizing high quality 3D shape will have a significant impact on various fields, ranging from entertainment to healthcare and metaverse. We believe our huge models can enabled 3D artists to design intricate structures while reducing costs and development time. Simultaneously, our focus is on advancing the development and application of larger models in the field of computer vision, particularly in the domain of 3D modeling. However, we are also cognizant of the need to address and mitigate the potential misuse of the powerful generative capabilities that these large models possess.

\paragraph{Limitation}
Although our large-scale models have shown improved performance in generating 3D shapes, their effectiveness and efficiency are constrained by the availability of training data and computational resources. In our future work, we plan to address this limitation by exploring the generation of high-quality 3D shapes as supplementary data using our model, which can be used to enhance the effectiveness of the first-stage training. Additionally, we will conduct experiments to investigate various methods and model architectures that can accelerate the inference process, aiming to improve the overall efficiency of our approach.

\end{document}